# Classification of Imbalanced Credit scoring data sets Based on Ensemble Method with the Weighted-Hybrid-Sampling

Xiaofan Liu[a], Zuoquan Zhang[a,*], Di Wang[a]

[a] School of Science, Beijing Jiaotong University, Beijing 100044, China

* Corresponding author. E-mail addresses: 16121655@bjtu.edu.cn (X. Liu), zuoquanzhang@163.com (Z. Zhang), 15118424@bjtu.edu.cn (D. Wang)

## Abstract

In the era of big data, the utilization of credit-scoring models to determine the credit risk of applicants accurately becomes a trend in the future. The conventional machine learning on credit scoring data sets tends to have poor classification for the minority class, which may bring huge commercial harm to banks. In order to classify imbalanced data sets, we propose a new ensemble algorithm, namely, Weighted-Hybrid-Sampling-Boost (WHSBoost). In data sampling, we process the imbalanced data sets with weights by the Weighted-SMOTE method and the Weighted-Under-Sampling method, and thus obtain a balanced training sample data set with equal weight. In ensemble algorithm, each time we train the base classifier, the balanced data set is given by the method above. In order to verify the applicability and robustness of the WHSBoost algorithm, we performed experiments on the simulation data sets, real benchmark data sets and real credit scoring data sets, comparing WHSBoost with SMOTE, SMOTEBoost and HSBoost based on SVM, BPNN, DT and KNN.

**Keywords**: Credit scoring, Imbalanced data, Data sampling, Boosting algorithm.

## 1 Introduction

The banks expect that more clients apply for their loans so as to get more profits. However, they have to face the risks that their applicants cannot pay them back. Therefore, it is very significant to give an accurate evaluation for those applicants' credit. Nowadays, credit scoring has become one of the primary ways of predicting the applicant's credit model, which is based on relevant attributes such as salary, employment, and credit history. In general, the models are developed to classify the applicants into two classes: good class and bad class. Applicants belonging to good class always have a larger possibility to repay loans and will be accepted by banks for loaning, while the opposite is true about those belonging to bad class. Actually, most applicants are more likely to pay off the loans within the stipulated time, which leads to the fact that the number of applicants belonging to good class is much larger than that of those belonging to bad class. As a result, credit scoring data sets are usually imbalanced data sets. How to identify the minority class effectively and accurately without raising the misjudgment of the majority is the key to improve the model.

In general, to overcome the imbalanced problem，the classifier has relied upon 3 broad approaches including feature selection, data processing, and algorithm-based approaches.

- Feature selection: if the irrelevant features are removed, the risk of discarding the minority will decrease

(Haixiang, 2016). Feature selection can improve the precision of classification effectively and reduce the complexity in time. Feature selection has become an important field in imbalanced data (Bae & Yoon, 2015; Haixiang, 2017; Wang, 2015), especially in high-dimensional applications (Maldonado, 2014; Moayedikia, 2017; Yin, 2013). It can be divided into filters, wrappers, embedded approach and hybrid method (Haixiang, 2017).

- Data processing: Sampling approaches aim to change the distribution of imbalanced data sets. It includes under-sampling approach by removing samples from the majority class and over-sampling approach by adding samples to the minority class. Random Under-sampling (Tahir, 2009), Condensed Nearest Neighbor Rule (CNN) (Hart, 1968), Neighborhood Cleaning Rule (NCL) (Laurikkala, 2001), One Sided Selection (OSS) (Kubat & Matwin, 1997), Tomek-Link (Tomek, 1976) and ENN (Wilson, 1972) are common under-sampling approaches. Most over-sampling approaches are limited to random sampling, the Synthetic Minority Over-Sampling Technique (SMOTE) (Chawla, 2002) method and SMOTE based generalizations (BSMOTE (Han, 2005), Safe-Level-SMOTE (Bunkhumpornpat, 2009), LN-SMOTE (Maciejewski & Stefanowski, 2011)). GarcaV studied the performance of the sampling method with different imbalanced ratios and classifiers, coming to the conclusion that over-sampling of the minority always outperforms under-sampling of the majority (García, 2012). C.Seiffert presented a hybrid sampling(HS) method (Seiffert, 2009) combining under-sampling and oversampling, which is widely applied. He proved that the hybrid sampling outperforms the individual methods for combining the strengths of the individual techniques and lessening the drawbacks.

- Algorithm- based approaches: The algorithm- based approaches improve the existing base classifiers so as to make the algorithm more sensitive to the minority class, but do not change the distribution of samples. The algorithm- based approaches can be divided into modifying single classifier (M'hamed & Fergani, 2014), cost-sensitive methods (Li, 2018; Cheng, 2017), and ensemble methods. Among them, ensemble methods can be strongly recommended especially for handling imbalanced problem, whose main idea is to generate a strong classifier by combining multiple base classifiers. Common ensemble learning (Soleymani, 2018; He, 2018) approaches include Boosting, Bagging, and their improvements. The development, application, advantages and disadvantages are discussed in detail in Section 2.

The strategies above solve the problem of imbalanced data from three different perspectives. In practice, however, a single strategy often fails to achieve good classification results. More and more scholars choose to use a hybrid of multiple strategies. The combination of sampling techniques and ensemble methods are common combination methods, such as SMOTEBOOST (Chawla, 2003), RUSBOOST (Seiffert, 2010), EUSBOOST (Galar, 2013), HSBoost (Lu, 2016), RB-BOOST (D'Addabbo & Maglietta, 2015).

In order to solve the problem of the imbalanced data, we propose two new sampling approaches based on boosting model for the first time, including the Weighted-SMOTE method and the Weighted-Under-sampling method. Thereby, a new ensemble method with the Weighted-Hybrid-Sampling called WHSBoost is established to solve the imbalanced classification. WHSBoost modified the weight-based resampling of training data set in the Adaboost algorithm. WHSBoost improves the diversity and weight utilization of base classifiers in ensemble method.

## 2 Ensemble Methods

### 2.1 Introduction to Ensemble Methods

Ensemble method is a machine learning algorithm that combines multiple single classifiers into a strong classifier according to a certain rule. The accuracy and diversity of the base classifiers are two important factors for ensemble methods. According to the types of the base classifiers, ensemble methods can be divided into homogenous ensemble methods and heterogeneous ensemble methods. The homogenous ensemble methods are to divide the training data set into several subsets and generate classifiers with different parameters by training with the same base classifier. Heterogeneous ensemble method refers to training with the same training data set and different base classifiers. Finally, several different base classifiers are eventually combined in heterogeneous ensemble methods.

According to the different training methods of the base classifiers, ensemble methods can be divided into parallel ensemble methods and sequential ensemble methods (Xia, 2017). The parallel method means that the training data set is divided into multiple subsets, and each corresponds to a base classifier. The common methods include bagging, random forest (RF) (Chen, 2004), and the multiple classifier systems (MCS) (Roli, 2015). During the construction of the classifier, the training subsets are generated in sequence, and the training subsets of the next base classifier are generated according to the previous results. The most commonly used method is the Adaboost algorithm. The AdaBoost algorithm has the advantages of simple structure, high accuracy and little influence by noise point, and is widely applied in credit scoring (Ghodselahi, 2011; Abellán & Castellano, 2017).

### 2.2 Research motivations

AdaBoost algorithm is one of the most commonly used ensemble methods. It increases the weight of the samples misclassified in the previous iteration and reduces the weight of the samples correctly classified in the previous iteration, so that the different base classifiers are obtained through different training sample data sets. The final classifier is integrated by weighted voting in AdaBoost algorithm.

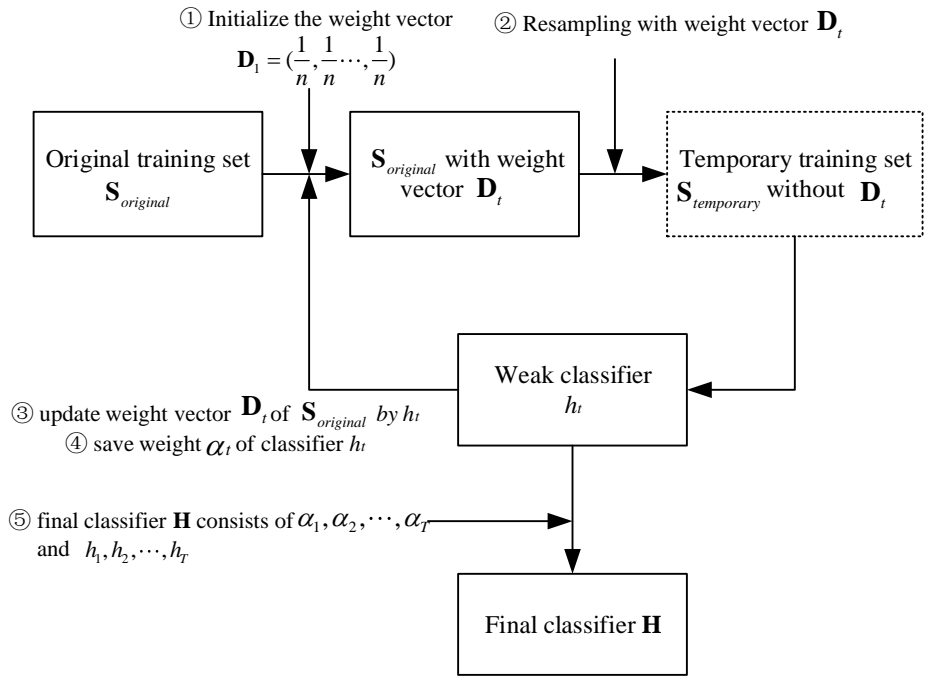

Fig.1 Flow diagram of AdaBoost

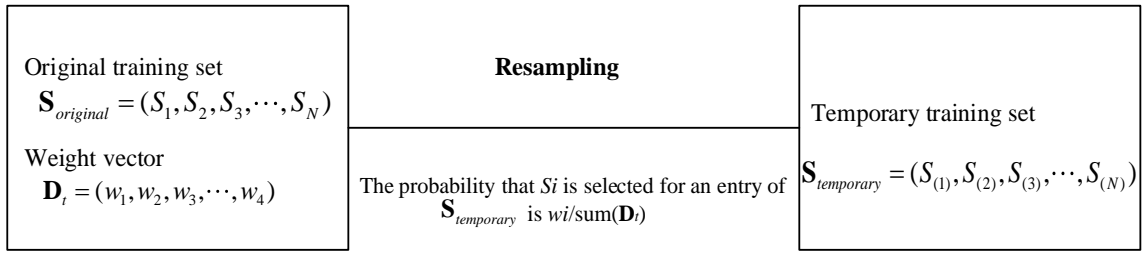

Fig.2 Flow diagram of the weight-based resampling

In AdaBoost algorithm, according to whether the base classifier can process weighted samples, the training process of base classifier can be divided into training by original weighted sample and training by resampling sample. As shown in Fig.1, SVM, DT, NN and other classifiers cannot deal with the weighted samples directly, so we need to resample data based on the weight and form a temporary training data set before the base classifier training. Fig.2 details the process of the weight-based resampling. In resampling, to ensure that the total number is constant after resampling, the majority samples are always selected by many times, while the minority is ignored. What's more, the above process ensures that the samples are distributed according to the weight and solves the problem that the base classifier cannot deal with the weighted samples. However, resampling leads to the reduction and repetition of training samples inevitably and increases the randomness of the algorithm. When the training data set is imbalanced or with small amounts, the reduction of the valid samples will lead to the decrease of the accuracy of the base classifier, thus reducing the overall accuracy.

At present, dealing with the imbalanced problem, there are mainly three improvements, as follows:

1) Improvements at the data level. During each iteration, we can get the temporary balanced training data set through

data processing techniques, and then use it to train the base classifier;

2) Optimization in the weight updates. Cost-sensitive ensemble algorithm is the most common. By introducing a cost factor to change the loss function for different categories of data, a new weight updating equation can be obtained. If we increase the misclassification cost of the minority class, the algorithm will pay more attention to the minority class in each iteration when generating base classifiers, so that the classification accuracy can be improved;

3) Replacement in the learning algorithm. Actually, Adaboost is an additive model. Loss function is an exponential function, and the learning algorithm is a two-step forward learning method. In order to enhance the generalization ability and calculating speed, we always make some learning algorithm improvements, such as extreme gradient boosting (XGBoost) (Xia, 2017). At the same time, combining with the improvements in data level is most widely used, which is effective and less difficult to operate.

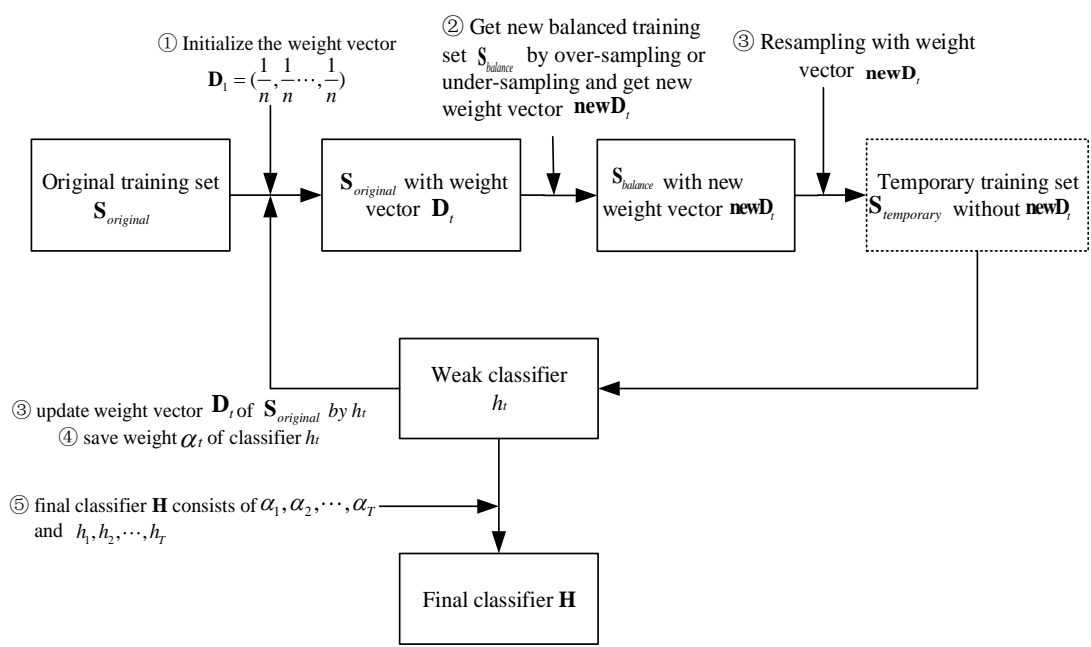

Fig.3 Flow diagram of AdaBoost based on data processing

Adaboost based on data processing, like RUSBoost, tends to ignore the weight-based resampling process, as shown in Fig.3, process 3. It makes the weight of samples underused and the diversity of base classifiers reduced. However, Adaboost based on data processing without ignoring the weight-based resampling in the process 3 of Fig.3, also results in damage of the weight distribution and large randomness of the algorithm.

At present, there is no literature to give a good improvement method of the process of the weight-based resampling for Adaboost algorithm. Accordingly, for the problem of imbalanced data sets, this paper proposes the Weighted-SMOTE method and the Weighted-Under-Sampling method instead of the weight-based resampling in Adaboost algorithm, and then we obtain WHSBoost algorithm. WHSBoost not only solves the problem of loss of sample information and ensures the full use of weight information, but also has good effects on imbalanced data sets.

# 3 WHSBoost: Ensemble method with the Weighted-Hybrid-Sampling

## 3.1 The Weighted-SMOTE method

SMOTE algorithm is the most representative oversampling method proposed by Chawla et al. in 2002 (Chawla, 2002; Douzas & Bacao, 2017). It is a method of synthesizing new data by samples and their nearest neighbor samples. On the one hand, it solves the problem of imbalanced data set, and on the other hand, it avoids the effect of overfitting. The characteristic attributes of the samples can be divided into continuous type and categorical type. SMOTE calculates the distance between two any continuous characteristic samples with Euclidean distance and calculates the distance between two categorical type samples with VDM distance.

In order to improve the classification of weighted imbalanced data set in ensemble method, in this paper, we firstly proposed the weighted SMOTE algorithm for weighted imbalanced samples. While increasing the number of the minority class, it ensures that the weight distribution of training data set remains unchanged, the sample diversity will be increased, and the weight will be fully utilized.

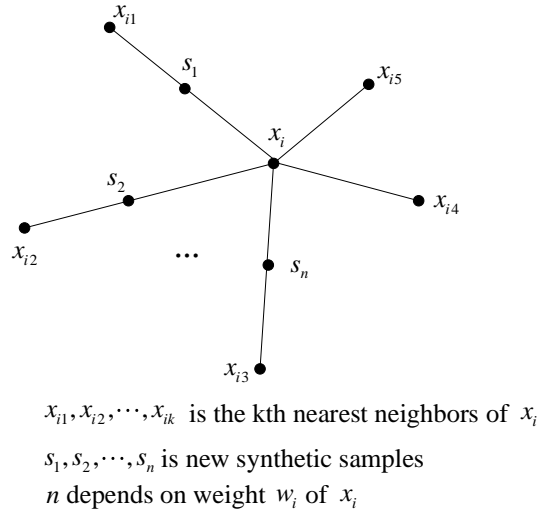

$x_{i1}, x_{i2}, \cdots, x_{ik}$ is the kth nearest neighbors of $x_i$

$s_1, s_2, \cdots, s_n$ is new synthetic samples

$n$ depends on weight $w_i$ of $x_i$

Fig.4 Sketch diagram of synthetic samples in WSMOTE

For base classifiers, the classification of weighted samples is equivalent to the classification of weight-resampled data set that is generated from original weighted samples. Therefore, it causes that some of the information is duplicated and some is wasted. The WSMOTE algorithm changes the distribution of the synthesized samples in the SMOTE algorithm and correlates them with the weight distribution of original samples. As shown in Fig.4, the number of synthesized samples, corresponding to each minority sample, is determined according to minority sample's weight. The approach ensures the new samples can be distributed based on original weight distribution and avoids excessive duplication or waste of sample information.

Algorithm 1 can be described as follows.

---

Algorithm 1 WSMOTE ( $\mathbf{T}, N, k, \mathbf{W}$ )

---

**Input**: minority class samples $\mathbf{T} = (\mathbf{T}_1, \cdots, \mathbf{T}_n) = \{(x_{11}, \cdots, x_{1m}), \cdots, (x_{n1}, \cdots, x_{nm})\}$, $N$ is the number of new samples, $k$ is the $k$th nearest neighbors, $\mathbf{W}$ is weight of minority class samples $\mathbf{W} = \{w_1, \cdots, w_n\}$, and $m$ is the number of feature attributes

1: **Procedure** WHSMOTE

2: Calculate the number of synthesized samples corresponding to each sample:

$$\mathbf{No} = (No_1, \cdots, No_n), \quad No_i = round(N * w_i), \text{ the } round \text{ indicates rounding;}$$

3: Check the total number of composite samples:

3.1: $\quad IX = sort(\mathbf{W}, 'descend')$, and IX indicates index in descending order of weight;

3.2: **If** $numNo = \sum_{i=1}^{n} No_i < N$:

$$\mathbf{No}(IX(1:numNo)) = \mathbf{No}(IX(1:numNo)) + 1;$$

3.3: **else if** $numNo > N$:

$$\mathbf{No}(IX(end - numNo + 1:end)) = \mathbf{No}(IX(end - numNo + 1:end)) - 1;$$

3.4: **end if**

4: Compute the $k$th nearest neighbors for each minority class sample:

4.1: **for** $i = 1, 2, \cdots, n$:

4.2: Calculate the distance between the $i$th sample and the $j$th sample:

$$Dis_j(T_i, T_j) = \sqrt{\sum_{ms}|x_{ims} - x_{jms}|^2 + \sum_{md} VDM(x_{imd}, x_{jmd})}$$

$j = 1, \cdots n$ and $j \neq i$, and $ms$ represents a continuous attribute, the distance is calculated by Mahalanobis distance, $md$ is categorical attribute, and the distance is calculated by VDM function;

4.3: Take $k$ samples of the smallest $Dis$ value as the $k$th nearest neighbors of the $i$th sample:

$$\mathbf{Ne}_i = (\mathbf{E}_1, \cdots, \mathbf{E}_k) = \{(e_{11}, \cdots, e_{1m}), \cdots, (e_{k1}, \cdots, e_{km})\};$$

4.4: **end for**

5: The synthesis process of new sample:

5.1: **while** $No_i \neq 0$:

5.2: for i, randomly select $No_i$ neighbors $(E_1, \cdots, E_{No_i})$ from $\mathbf{Ne}_i$, composite the new sample $\mathbf{Syn}_i = (\mathbf{S}_1, \cdots, \mathbf{S}_{No_i})$

5.3: Calculate the composite sample:

$$S_j = (s_{j1}, \cdots, s_{jm}), j = 1, \cdots, No_i$$

for the continuous attributes of $S_j$, $s_{jms} = x_{ims} + r_j * (x_{ims} - e_{jms})$, $r_j$ is a random number from 0 to 1; For categorical attributes of $S_j$, $s_{jmd}$ is voted by $e_{1md}$ to $e_{No_i md}$.

5.4: **end while**

6: **End procedure**

**Output:** The synthetic sample is $\mathbf{Syn} = (\mathbf{Syn}_1, \cdots, \mathbf{Syn}_n)$, new sample set of minority class is $\mathbf{T}_{new} = \mathbf{T} \cup \mathbf{Syn}$

Algorithm 1 In Fig.4, firstly, we use the weight of minority samples to determine the number of synthesized samples corresponding to each sample. Then the K-nearest neighbor of each sample can be obtained. Finally, synthesized samples are generated by original sample and its nearest neighbor randomly. It is noted that the continuous features and categorical features are different in calculating the distance and generating the synthetic samples. In this paper, the biggest difference between WSOMTE and SMOTE is how to determine the number of synthesized samples. The SMOTE algorithm treats each minority sample as equal, and distributes the synthesized samples evenly, but cannot process weighted samples. The WSMOTE algorithm, however, changes the distribution of the synthesized samples so that the new sample $\mathbf{T}_{new}$ can distribute similarly to the original samples essentially. When training the classifier, the function of the original weight can be realized by the new sample $\mathbf{T}_{new}$. At the same time, by WSMOTE algorithm, the number and the diversity of samples are increased, and the classifier training is more beneficial.

3.2 The Weighted-Under-Sampling method

Under-sampling is a sampling method that reduces the imbalanced ratio by applying certain techniques to reduce the size of the majority. The simplest under-sampling method is to remove some majority samples randomly. This method is able to balance the imbalanced data set and is easy to operate. However, some useful information may also be removed at the same time as the majority class is removed, and some samples with strong discriminative properties may be lost. In order to make up for this, we elicit the sample weight to reflect the sample importance. The weight is determined based on the Adaboost process, which aims to improve the accuracies of base classifiers. Algorithm 2 can be described as follows.

Algorithm 2 WUSample ( $\mathbf{T}$ , $N$, $\mathbf{W}$ )

**Input**: majority class samples $\mathbf{T} = (\mathbf{T}_1, \cdots, \mathbf{T}_n)$ ; $N$=number of new samples; weight of majority class samples W

1: **procedure** WUSample

2: Sort weights in ascending order:

$IX = sort(\mathbf{W})$ , and IX indicates sample index after sorting;

3. Construct a random sample set: Take the weights in ascending order in the top $N+c*N$ , $N+c*N \leq n$ sample, and form a sample set $\mathbf{S}$ , where c is a constant from 0 to 1;

4: Select N samples randomly from sampling data set A and constitute the elimination sample set $\mathbf{D}$ ;

5: New sample set of majority class is $\mathbf{T}_{new}$ : $\mathbf{T}_{new} = \mathbf{T} \cap \mathbf{D}$

6: **end procedure**

**Output:** New sample set of majority class is $\mathbf{T}_{new}$

Algorithm 2 Firstly, we construct the random sampling data set $\mathbf{S}$ according to the weights, and then extract the eliminated samples from the sampling data set randomly. The number of samples in a random sampling data set $\mathbf{S}$ is generally greater than or equal to N. $c$ is a constant controlling the number of samples in $\mathbf{S}$ , which reflects the randomness of the WUSample algorithm. When $N+c*N=n$, WUSample algorithm is equivalent to the under-sampling algorithm.

### 3.3 WHSBoost

Under-sampling results in the loss of data information and over-sampling alone will lead to over-fitting in the training of base classifiers. Hybrid sampling combines both methods effectively and gets balanced data sets. Therefore, in order to solve the imbalance, this paper proposes the ensemble method with the Weighted-Hybrid-Sampling called WHSBoost.

The main idea of WHSBoost is to use the Weighted-Hybrid-Reampling method combined with the Weighted-WSOMTE and the Weighted-Under-Sampling, replacing the weight-based resampling in Adaboost. As for the data processing based on the Adaboost algorithm (Fig.3, process 2) and the process of the weight-based resampling (Fig.3, process 3), both are replaced by the Weighted- Hybrid-Reampling in the WHBOOST algorithm. The method simplifies the algorithm and reduces the loss and reuse of sample information.

---

**Algorithm 3 WHSBOOST ($\mathbf{S}_{original}$, $L$, $T$, $N$, $k$)**

**Input**: The original training data set $\mathbf{S}_{original} = \{(\mathbf{x_1}, y_1), \cdots, (\mathbf{x_n}, y_n)\}$, $L$ = Base classifier, $T$ = the number of iterations in boosting procedure, $N$ = The number of the majority samples when the data is balanced (the minority), $k$ is the $k$th nearest neighbors

1: **procedure** WHSBoost

2: Initialize the weight vector:
$$\mathbf{D}_1 = \{D_1(1), \cdots, D_1(n)\}, \quad D_1(i) = 1/n, \quad for\ i = 1, 2, \cdots, n;$$

3: **for** $t = 1, 2, \cdots T$:

4:     Get the temporary balanced training data set $\mathbf{S}_t$:

4.1:     $\mathbf{SA}_{original}$ is the majority class of $\mathbf{S}_{original}$, the weight vector of $\mathbf{SA}_{original}$ is $\mathbf{DA}_t$, then
$$\mathbf{SA}_t\ (N\ \text{samples}) = \text{WSMOTE}(\mathbf{SA}_{original}, N, k, \mathbf{DA}_t);$$

4.2:     $\mathbf{SI}_{original}$ is the minority class of $\mathbf{S}_{original}$ and the weight vector of $\mathbf{SI}_{original}$ is $\mathbf{DI}_t$, then
$$\mathbf{SI}_t\ (N\ \text{samples}) = \text{WUSample}\ (\mathbf{SI}_{original}, N, \mathbf{DI}_t)$$

4.3:     the temporary balanced training data set $\mathbf{S}_t = \mathbf{SA}_t \cup \mathbf{SI}_t$

5:     Train a week classifier $L_t$ with $\mathbf{S}_t$ and calculate a weak hypothesis $h_t : \mathbf{x} \to y$

6:     Calculate the pseudo-loss based on set $\mathbf{S}_t$:
$$\varepsilon_t = \sum_{i=1}^{n} D_t(i) \mathrm{I}(h_t(\mathbf{x_i}) \neq y_i) / \sum_{i=1}^{n} D_t(i)$$
where I is an indicator function and $(\mathbf{x_i}, y_i) \in \mathbf{S}_{original}$

7:     Calculate the weight of the basic classifier:
$$\alpha_t = \frac{1}{2} \ln \frac{1 - \varepsilon_t}{\varepsilon_t}$$

8: Update the weight:

$$D_{t+1}(i) = D_t(i) \cdot \begin{cases} \exp(\alpha_t) & \text{if } h_t(\mathbf{x_i}) \neq y_i \\ 1 & \text{if } h_t(\mathbf{x_i}) = y_i \end{cases}, (\mathbf{x_i}, y_i) \in \mathbf{S}_{original}$$

$$D_{t+1}(i) = \frac{D_{t+1}(i)}{\sum_{i=1}^{n} D_{t+1}(i)}$$

9: **end for**

10: **end procedure**

**Output:** The final classifier $H(x) = \text{sign}(\sum_{t=1}^{T} \alpha_t \cdot h_t(\mathbf{x}))$

Algorithm 3 can be described as follows. First, we initialize the weight matrix of the original samples, which will be updated in the weight update process. In each iteration of WHSBOOST, the combination method of the Weighted-SMOTE and the Weighted-Under-Sampling method will generate the temporary balanced data set. It means to make the number of the majority class equal to that of the minority class, which is different from the step of other algorithms such as HSBOOST. The HSBOOST algorithm uses the hybrid sampling to obtain the balanced data set first, which results in the scatter of the minority class and the concentration of the majority. Subsequently, the weight-based resampling process increases the probability of the majority being drawn, and the resulting temporary data set becomes imbalanced again. However, the WHSBOOST algorithm avoids that. It guarantees that $S_{original}$ is the balanced data set in each iteration, and the weight distribution is almost the same as $S_{original}$. The temporary balanced data set is trained by the base classifier $h_t$. According to the performance of the base classifier on $S_{original}$, the weight and the weight parameter $\alpha_t$ of the base classifier are calculated. When the base classifier reaches the maximum iteration times $T$, or the error rate of the classification is less than the threshold, stop the iteration. The final strong classifier $H(x)$ consists of each base classifier $h_t(x)$ and its weight parameter $\alpha_t$. Fig.5 also graphically represents the entire process of WHSBoost.

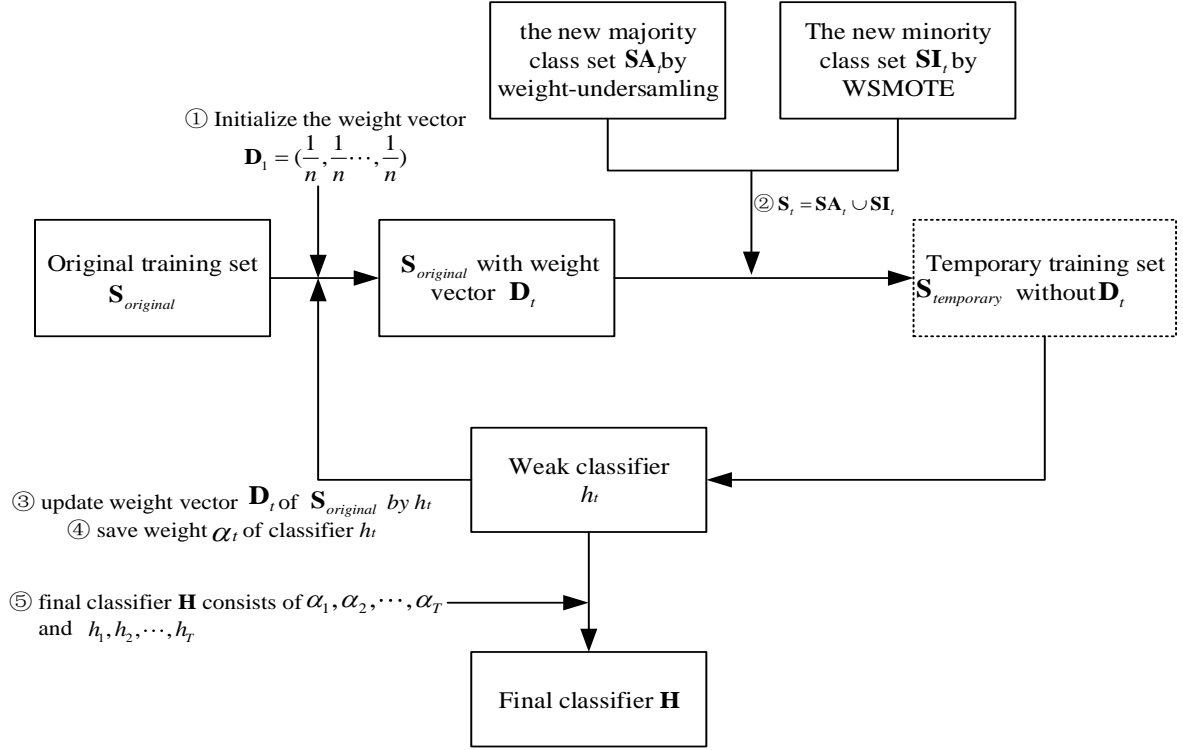

Fig.5 Flow diagram of WHBOOST

## 4 Experiment

### 4.1 Base classifiers

In this paper, we choose Support vector machines (SVM), Back Propagation neural networks (BPNN), Kth Nearest Neighbor (KNN), and Decision tree (DT) as base classifiers. These methods have been widely used in classification problems, with fast computing speed and good stability, which can solve the credit scoring problem effectively and quickly.

#### 4.1.1 Support vector machines

SVM is a machine learning method based on statistical theory and structural risk minimization principle proposed by Vapnik in 1995, whose strength lies with its special ability of dealing with small samples, nonlinear and multidimensional pattern recognition (Ala'raj & Abbod, 2016). For classification problems, its main issue is how to build a classification hyperplane as the decision plane. For imbalanced data sets, the larger penalty parameter C is, the bigger the mistakes are allowed for the minority, and then the decision plane will be easier toward the minority class. As a result, the prediction may lead to a high error rate in the minority class. In this paper, we choose Gaussian RBF Kernel for kernel function for credit scoring data sets.

#### 4.1.2 Back Propagation neural networks

Neural network is a kind of nonlinear simulation technology. BP neural network is a kind of multilayer feed forward neural network proposed by Rumelhart and MeClelland in 1986, whose main characteristic is error back propagation. A complete BP neural network consists of input layer, hidden layer and output layer (Ala'raj & Abbod,

2016). In the process of forward spread, the input signal is layer by layer processed from the input layer to the hidden layer and finally output from the output layer. In fact, the input and output of BP neural network are respectively the independent variables and the dependent variables of the function.

### 4.1.3 Kth nearest Neighbor

Kth Nearest Neighbor classification algorithm is one of the most common machine learning algorithms. The main idea is that if a sample has most of the *k* nearest samples belonging to a category in the feature space, the sample also belongs to this category (Bequé & Lessmann, 2016). The KNN algorithm does not perform well on the imbalanced problem, as the minority sample is easily misclassified. The k-nearest neighbor of any new sample will occupy the majority, which has a strong impact on the classification result.

### 4.1.4 Decision tree

The Decision Tree proposed by Quinlan in 1979 is a tree construction similar to flow chart, consisting of node and directed edge. Each internal node represents a test of a feature, each branch represents an output of the test, and each leaf node represents a class label (Ala'raj & Abbod, 2016). Decision Tree starts from the unique root node of the tree and uses a top-down recursive method. ID3 and C4.5 respectively calculate the purity using the information gain and the information ratio. In this paper, we choose C4.5.

### 4.2 Classification Evaluation Criteria

The different evaluation criteria of classification mean different focus on the classification results, so the selection of evaluation criteria plays an important role in the final evaluation result. For imbalanced data sets, Recall, F-Score, and Area Under Curve (AUC) are often utilized as evaluation criteria (Haixiang, 2017; Abdou & Pointon, 2011). In this paper, we classify the minority as the positive class and the majority as the negative class, and the confusion matrix as shown in Table 1.

Table 1 Dichotomous confusion matrix

|  | Forecast is positive | Prediction is negative |
|---|---|---|
| Positive samples | True Positive(TP) | False Negatives(FN) |
| Negative samples | False Positive(FP) | True Negatives(TN) |

Table 2 shows the relevant evaluation metrics.

Table 2 Dichotomous assessment metrics

| Assessment | Accuracy | Recall | Precision | F-Score |
|---|---|---|---|---|
| Formula | $\dfrac{TP+TN}{TP+FP+TN+FN}$ | $\dfrac{TP}{TP+FN}$ | $\dfrac{TP}{TP+FP}$ | $\dfrac{(1+\beta^2)*recall*precision}{\beta^2*precision+recall}$ |

For imbalanced data sets, the accuracy cannot really reflect the performance of the classifier. In credit scoring, the prediction accuracy of the minority class is as important as the majority class, but the accuracy is likely to generate deceptive high accuracy. Recall reflects the proportion of the positive (minority) samples being predicted correctly. The higher the ratio, the stronger the ability to correctly identify the minority samples is. Precision reflects the possibility of misclassification of negative (majority) samples. F-score is a weighted average of the Recall and precision, and the larger the score, the better the overall production result. The greater the value of $\beta$, the greater the proportion of Recall is. In order to better evaluate the prediction accuracy of minority class, we take $\beta=3$.

Credit scoring system is a typical problem of classifying imbalanced data sets. The prediction accuracies for both positive and negative classes ought to be as high as possible in order to ensure the quality of applicant sources. Therefore, this paper uses Recall and F-Score to score the comprehensive evaluation model (Abdou & Pointon, 2011). In addition, the AUC value is also used as an index to evaluate the model. The higher the value, the better the classifier is. When 0.5 <AUC <1, the model is better than the random guess (He, 2018).

### 4.3 Experiments on different imbalanced ratios

In this paper, the applicability and robustness of the WHSBoost algorithm will be verified by training imbalanced data sets with different imbalanced ratios by numerical simulation and benchmark data sets. Simulation data sets are classified based on linear hyperplane. 20 benchmark data sets in different fields are nonlinear data sets, similar to the credit scoring data sets. In order to show the advantage of WHSBoost algorithm, we compare SMOTE, SMOTEBoost, HSBoost and WHSBoost based on SVM, BPNN, KNN and DT. The HSBoost algorithm is a combination of hybrid sampling and the ensemble algorithm under normal circumstances, as shown in Fig.3.

#### 4.3.1 Simulation

In the simulation, we generated a total of 1000 samples $(\mathbf{x}_i, y_i)$. The total number of features is $p = 25$, and the number of effective features is $p_0 = 8$. Each feature has the same distribution as $N(0,1)$, then $x_1, x_2, \cdots, x_p$ are independently sampled from the standard normal distribution. Then, we extract $p_0$ from the $p$ total features to construct the classifying hyperplane $\{\mathbf{x} \mid g(x) = \beta_1 x_{(1)} + \cdots + \beta_{p_0} x_{(p_0)}\}$, and obtain the sample label $y_i = \text{sgn}(g(\mathbf{x}_i))$ (Gong & Kim, 2017). In order to construct samples with different ratios, firstly, 5000 instances are generated at a time, only 1000 of them are chosen in order to adjust the degree of imbalance in the data set. Considering common imbalanced ratios in the credit scoring model, the ratios of the minority class are set to be 45%, 30%, 20% and 10%. In order to study the performance of simulation data sets, the 1000 samples are divided into two random parts with the same imbalanced ratio, 20% of instances as tuning part and the rest 80% as performance estimation part. In the tuning part of the data, hyper-parameters of classifier are determined. The performance estimation part randomly divided 100 times, of which 70% were training data and 30% were test data.

In order to select the optimal hyper-parameters, we use PSO algorithm to optimize the penalty parameter C and parameters of the kernel function for SVM, use grid search algorithm to optimize the number of hidden layer neurons in BPNN and the number of neighbors of KNN classifier. The flow of parameter optimization method is as follows. 1) Randomly divide tuning part of the data, select 80% samples as the tuning train set and 20% samples as the tuning verification set. 2) SMOTE algorithm is used to process the tuning train set to get balanced data set. 3) Use PSO algorithm or grid search algorithm for parameter optimization. When PSO algorithm is used, the optimization objective function is the loss of tuning verification data set. When grid search algorithm is used, the evaluation criterion is the accuracy of verification data set. For simulation data sets, we do not normalize the data. When the imbalanced ratio is 0.1, 0.2, 0.3 and 0.45, the linear kernel of SVM is taken, and the penalty parameter C is 2.062, 2.4, 4.961 and 1.751, respectively. The number of hidden layer neurons in BPNN classifier is 8, the number of iterations is 20, and the learning rate is 0.1. The number of neighbors K of KNN classifier is 3, 3, 5 and 5 respectively. The maximum depth of DT classifier is 10.

We choose the F-score, Recall, AUC value as evaluation criteria. The means of evaluation index results on SVM, BPNN, KNN, DT are shown in Appendix I, Table 1-4.

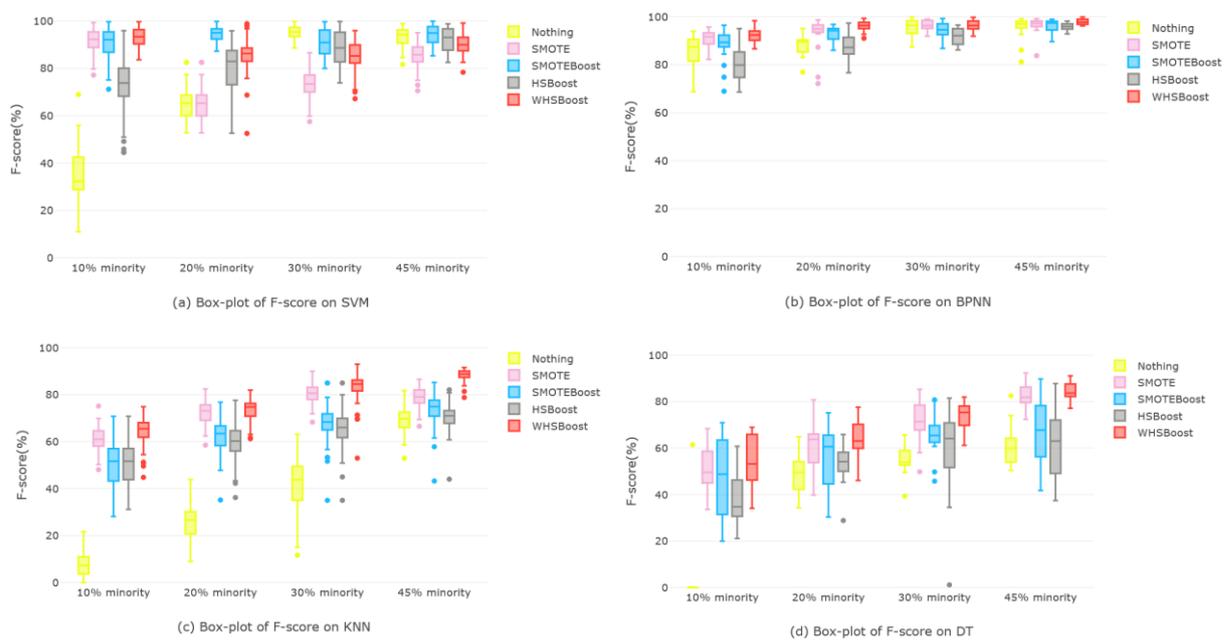

Fig.6 Box-plot of F-score on SVM, BPNN, KNN and DT

As is shown in Fig.6 and Appendix I, among five types of data processing algorithms dealing with data sets with different classifiers and different imbalanced ratios, The F-score, recall and AUC of WHSBoost are superior to others in most cases.

We compare WHSBoost algorithm with two types of algorithms, traditional data sampling algorithms and algorithms combining data sampling and ensemble algorithm. SMOTE algorithm is the most commonly used traditional data sampling method. For linear data sets with different imbalanced ratios, the accuracy and stability of SMOTE rank second, only inferior to WHSBoost. However, when the base classifier has poor classification ability (such as KNN and DT), the accuracy of SMOTE is slightly lower than that of WHSBoost. On the one hand, WHSBoost solves the problem of imbalanced data sets and improves the accuracy of single base classifier. On the other hand, WHSBoost combines base classifiers into strong classifiers, which improves the overall classification effect. SMOTEBoost and HSBoost are combined models of traditional data sampling and ensemble algorithms, as shown in Fig.3. In most cases, the stability and accuracy of SMOTEBoost and HSBoost are lower than SMOTE and WHSBoost. The main reasons are as follows: (1) SMOTEBoost and HSBoost resample the training data set with sample weight distribution for a single base classifier. According to Fig.2, the weight-based resampling process of weighted samples will lead to imbalance in training data sets, and as a result, the classification accuracy of a single base classifier does not improve obviously, even lower than that of traditional data sampling algorithms such as SMOTE. Moreover, the weight-based resampling process reduces the stability of the base classifier. (2) Ensemble algorithm is a method that combines multiple base classifiers into strong classifiers. When the classification ability of a single base classifier is poor, the classification accuracy of base classifier will be improved by ensemble algorithm, but it is still slightly worse than that of other classifiers improved by same ensemble algorithm. As shown in Fig.6, the accuracies of KNN and DT base classifiers are low. The accuracies of SMOTEBoost-KNN, HSBoost-KNN, SMOTEBoost-DT and HSBoost-DT are lower than other algorithms, and the variances are relatively large. For the SVM and BPNN base classifiers, the accuracy of SMOTEBoost is similar to that of SMOTE and WHSBoost, especially when the imbalanced ratio is 30% and 45%. (3) HSBoost algorithm is a combination of SMOTE algorithm and complete random sampling algorithm. Random sampling will result in strong randomness in the samples and thus lead to large fluctuations. WHSBoost algorithm improves HSBoost aiming at its disadvantages, which not only avoids the negative effects of weight-based resampling process, but also reduces the randomness of the algorithm. Therefore, WHSBoost is obviously superior to other ensemble algorithms.

From the view of different base classifiers, WHSBoost performs better in BPNN, while KNN and DT perform poorly. In terms of the basic principles of models, the complexity of BPNN classifiers is evidently higher than other base classifiers, especially KNN and DT. Moreover, the accuracy of DT classifier is lower than that of SVM and BPNN because the features of the simulation data sets are continuous, independent and identically distributed. Although compared with SMOTEBoost and HSBoost, WHSBoost is less affected by base classifiers, base classifiers still have some effects on the accuracy of WHSBoost. We need to select an appropriate base classifier in practical application.

4.3.2 Benchmark data sets

The 20 benchmark data sets are derived from KEEL repository (https://sci2s.ugr.es/keel/imbalanced.php), and the imbalanced ratio is between 1.86 and 39.15, seeing Table 5 of Appendix I. The results of 20 benchmark data sets are shown in Fig.1, Fig.2, and Fig.3 of Appendix I.

We can see that Recall, F-score and AUC fiercely fluctuate in different data sets on base classifiers. The Recalls of some data sets on base classifiers are 0, which further proves that base classifiers cannot deal with imbalanced data sets effectively. As is shown in Fig.3 of Appendix I, the minimum value of Recall of WHSBoost is around 60%, and the median value is above 85%, and the range is obviously smaller than other algorithms. As is shown in Fig.2 of Appendix I, the minimum value of F-score of WHSBoost is about 60%, and the median is about 90%. The results vary weakly on different classifiers. It can be seen that WHSBoost is suitable for data sets with different imbalanced ratios and improves the accuracies of different base classifiers. In summary, WHSBoost not only has a good classification effect on linear data sets, but also applies to nonlinear cases.

### 4.4 Experiment on real credit scoring data sets

#### 4.4.1 Data Preprocessing

In order to verify the effect of WHSBoost algorithm in real credit-scoring, 8 credit scoring data sets were used to compare the performance of different data processing algorithms and different base classifiers. Table 3 is a brief description of credit scoring data sets. The Australian, German, Japanese, Credit approval, DefaultData credit scoring data sets are obtained from the UCI repository of machine learning (https://archive.ics.uci.edu/ml/data sets.html). The QHaidata_A, QHaidata_B credit scoring data sets are obtained from a Chinese credit consulting company named QianHai credit consulting company (https://www.kesci.com). QHaidata_A data set is user data of bank loans, and QHaidata_B data set is user data of Payday loans. LC2018Q1Data data set contains loan data of the first quarter in 2018 from Lending Club (https://www.lendingclub.com). The details of the data sets are showed in Table 3.

Table 3 credit scoring data sets

| Name | N | Features | Good sample | bad sample |
|---|---|---|---|---|
| German | 1000 | 24 | 700 | 300 |
| Australian | 690 | 14 | 307 | 383 |
| Japanese | 653 | 15 | 296 | 357 |
| Credit approval | 690 | 15 | 307 | 383 |
| DefaultData | 30000 | 23 | 23364 | 6636 |
| QHaidata_A | 4000 | 20 | 3487 | 513 |
| QHaidata_B | 4000 | 25 | 3719 | 281 |
| LC2018Q1Data | 107865 | 43 | 107812 | 53 |

In Table 3, Column "N" is the number of samples in the data sets. Column "Features" is the number of features. The features of each data set include both numeric and nominal ones. To ensure the relatively high accuracy, we eliminate the features with more vacancies in QHaidata_A, QHaidata_B, LC2018Q1Data data sets. We generally define the minority as the positive class. For hyper-parameters selection, each scoring data set is divided into two random parts, 20% of instances as tuning part and the rest 80% as performance estimation part. In the tuning part of the data, hyper-parameters of classifier are determined. In order to verify the classification effects of each model, we selected 70% of performance estimation part in each credit data set randomly as the training samples, and the rest as the testing samples. To observe the impact of randomness on the classification in data sampling, model training process is repeated 50 times and the mean of the results is taken.

#### 4.4.2 Experimental Results

In the real data sets, the subtle differences in classification can result in huge economic benefits. In the same way as the simulation data sets, we compare WHSBoost algorithm with SMOTE, SMOTEBoost and HSBoost based on SVM, BPNN, KNN and DT. Table 4 shows the AUC of classification on credit scoring data sets, and the F-score and the Recall of 8 credit scoring data sets are shown in Table 1 and Table 2 of Appendix Ⅱ.

Table 4 AUC of classification on credit scoring data sets

|  |  | German | Australian | Japanese | Credit approval | Default Data | QHaidata_A | QHaidata_B | LC2018Q1Data |
|---|---|---|---|---|---|---|---|---|---|
| SVM | Nothing | 0.792 | 0.917 | 0.894 | 0.530 | 0.587 | 0.514 | 0.519 | 0.573 |
|  | SMOTE | 0.777 | **0.917** | 0.896 | 0.748 | 0.576 | 0.526 | 0.558 | 0.585 |
|  | SMOTEBoost | 0.508 | 0.651 | 0.649 | 0.723 | 0.505 | 0.505 | 0.523 | 0.547 |
|  | HSBoost | 0.518 | 0.554 | 0.527 | 0.555 | 0.561 | 0.532 | 0.501 | 0.550 |
|  | WHSBoost | **0.790** | 0.913 | **0.910** | **0.760** | **0.592** | **0.535** | **0.558** | **0.603** |
| BPNN | Nothing | 0.728 | 0.900 | 0.885 | 0.654 | 0.568 | 0.513 | 0.538 | 0.532 |
|  | SMOTE | 0.705 | 0.892 | 0.860 | 0.687 | 0.592 | 0.537 | 0.523 | 0.586 |
|  | SMOTEBoost | 0.760 | 0.908 | 0.893 | 0.682 | 0.573 | 0.541 | 0.532 | **0.631** |
|  | HSBoost | 0.763 | 0.909 | 0.893 | 0.682 | 0.570 | 0.523 | 0.525 | 0.573 |
|  | WHSBoost | **0.767** | **0.910** | **0.901** | **0.691** | **0.648** | **0.547** | **0.553** | 0.624 |
| KNN | Nothing | 0.703 | 0.800 | 0.896 | 0.661 | 0.612 | 0.522 | 0.540 | 0.495 |
|  | SMOTE | 0.738 | 0.897 | 0.895 | 0.735 | 0.596 | 0.549 | **0.560** | 0.595 |
|  | SMOTEBoost | 0.710 | 0.856 | 0.857 | 0.718 | **0.625** | 0.536 | 0.517 | 0.567 |
|  | HSBoost | 0.701 | 0.850 | 0.860 | 0.726 | 0.617 | 0.551 | 0.533 | 0.574 |
|  | WHSBoost | **0.791** | **0.889** | **0.898** | **0.759** | 0.612 | **0.588** | 0.552 | **0.598** |
| DT | Nothing | 0.500 | 0.676 | 0.676 | 0.500 | 0.500 | 0.500 | 0.500 | 0.500 |
|  | SMOTE | 0.579 | 0.676 | 0.675 | 0.539 | 0.550 | 0.507 | 0.526 | 0.508 |
|  | SMOTEBoost | 0.577 | 0.717 | 0.730 | 0.556 | 0.569 | 0.535 | 0.500 | 0.524 |
|  | HSBoost | 0.543 | 0.719 | 0.676 | 0.553 | 0.542 | 0.482 | 0.488 | 0.473 |
|  | WHSBoost | **0.589** | **0.725** | **0.739** | **0.624** | **0.570** | **0.615** | **0.552** | 0.572 |

For different data sets, we use PSO optimization algorithm and grid search algorithm to get the hyper-parameters of different classifiers. In order to improve the accuracy, we normalize the data. For the SVM base classifier, we choose the Gauss kernel function with penalty parameter C of 23.42, 3.5, 1.63, 19.183, 20.452, 10.459, 4.03, 2.194 and gamma of 0.001, 2.043, 1.358, 3.59, 2.306, 2.101, 0.897 and 1.149, respectively. For the BPNN base classifier, the number of iterations is 20, and the number of hidden layer neurons is 10, 5, 5, 5, 10, 10, 15 and 15, respectively. The number of neighbors K of KNN classifier is 3, 5, 5, 5, 8, 5, 5 and 3, respectively.

The AUC value reflects the stability of the algorithm, which is a comprehensive criterion to evaluate the performance of classifiers. Table 4 shows that for most credit scoring data sets, the AUC of WHSBoost is the highest. On LC2018Q1Data data set, the AUC of SMOTE-BPNN is 0.631, which is higher than that of WHSBoost-BPNN, but the F-score of WHSBoost-BPNN is 51.109%, higher than that of SMOTE-BPNN. In short, for credit scoring data sets, the classification only induced by base classifiers cannot meet the requirements of the economy and society. WHSBoost algorithm has the highest accuracy, followed by SMOTE algorithm.

From Table 4, we can see that the selection of classifiers is very important to the results. For the SVM and KNN base classifiers, the differences of AUC among different algorithms are large. It is mainly caused by the fact that the base classifier is sensitive to the imbalance in data set and the accuracy is greatly affected by the training data sets. The accuracy of DT classifier is slightly lower than that of other algorithms, and SVM classifier is the most suitable for the classification of credit scoring data sets.

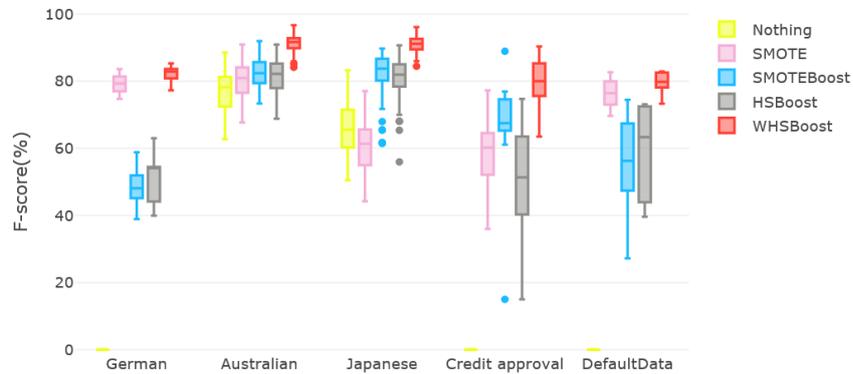

Fig.7 F-score of German, Australian, Japanese, Credit approval, DefaultData data sets on SVM

German, Australian, Japanese, Credit approval, DefaultData credit scoring data sets are all from UCI database. Among them, the number of samples, the imbalanced ratios and the number of features are similar. The imbalanced ratios of German data sets and DefaultData data sets are relatively large, and the accuracies are lower than other data sets. The F-scores on SVM of 5 credit scoring data sets are shown in Fig.7, and their F-scores on BPNN, KNN and DT are displayed in Fig.1, Fig.2, and Fig.3 of Appendix Ⅱ.

Compared with box-plots of F-scores of different base classifiers, SVM is a suitable base classifier for credit scoring problems. The accuracy of SVM and BPNN is higher than that of KNN, especially in DefaultData credit scoring data sets. The stability of DT and BPNN is poor, and the variance of F-score on different data sets is higher than that of SVM. Although under the cases when the SVM based classifiers are only used for classification, the F-scores of German, Credit approval, and DefaultData credit scoring data sets are all 0, SMOTE and WHSBoost algorithms improve the accuracy of SVM obviously, which is higher than other base classifiers.

Fig.7 shows that the F-score of WHSBoost-SVM is averagely 61.609% higher than that of SVM, and 10% higher than that of the other 3 data processing algorithms. There are obvious advantages in classifying the credit scoring data sets. At the same time, the stability of WHSBoost-SVM is better than other algorithms, especially traditional ensemble algorithms such as SMOTEBoost and HSBoost. On German, Credit approval, DefaultData credit scoring data sets, the SVM performs poorly, and F-score is 0. The accuracies of SMOTEBoost-SVM and HSBoost-SVM is obviously worse, and the F-score fluctuates more widely. The accuracies of SMOTEBoost-SVM and HSBoost-SVM for Australian and Japanese are higher than that of SMOTE. Thus we can get the conclusion that SMOTEBoost and HSBoost are strongly influenced by base classifiers, which is the same as the conclusion in Section 4.3.

In a word, the classification results of 5 credit scoring data sets form UCI verify the validity of WHSBoost proposed in this paper. It can not only accurately identify the "bad samples" in credit scoring problems, but also reduce the misjudgment of "good samples". It also proves that WHSBoost algorithm is not only suitable for cases with small sample, but also has a stable classification effect for cases with large sample. At the same time, the stability of WHSBoost is satisfactory, which can meet the requirements of credit scoring and maximize economic benefits.

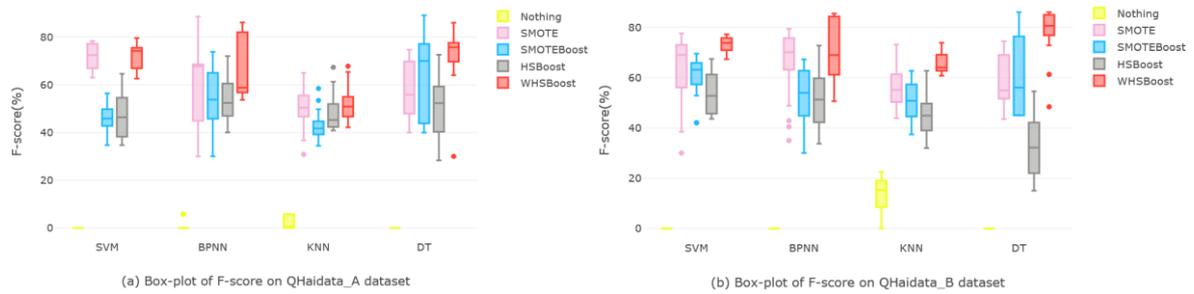

Fig.8 Box-plot of F-score on QHaidata_A data set and QHaidata_B

5 credit scoring data sets from UCI database are data sets mainly for traditional credit problems. In recent years, China's credit market has developed rapidly. Although it started later than other countries, the market has expanded rapidly and there are many types of loans, especially Payday Loans. China's economic system has also led to a different credit scoring problem from other countries. QHaidata_A and QHaidata_B data sets are obtained from a Chinese credit consulting company named QianHai. QHaidata_A data set is user data of bank loans, and QHaidata_B data set is user data of Payday loans. Therefore, these two data sets are pretty meaningful in verifying whether WHSBoost algorithm can effectively do with credit scoring problems.

As can be seen from Fig.8, the classification results of WHSBoost algorithms on QHaidata_A are similar to those on DefaultData data set, which belongs to the first echelon. The classification results of WHSBoost on QHaidata_B accord with the exceptions as well, which proves that WHSBoost is capable of dealing with credit-scoring in Payday loans.

For the SVM base classifier, the classification results of WHSBoost-SVM algorithm are similar to those of SMOTE-SVM algorithms. However, the F-score distribution of WHSBoost algorithm is more concentrated and the fluctuation is smaller. On QHaidata_A data set, except WHSBoost algorithm, the minimums of F-scores of the 5 algorithms is about 35%. On QHaidata_B data set, the minimums of F-scores of HSBoost-DT algorithms are even less than 20%. The F-score fluctuation of WHSBoost algorithm on QHaidata_B data set is less than half of the average fluctuation of the other algorithms. The accuracies of WHSBoost on Payday loans and bank loans are similar, and WHSBoost possesses higher accuracy and stability on different base classifiers.

In conclusion, the results of QHaidata_A and QHaidata_B data sets prove the validity and stability of WHSBoost algorithm in dealing with credit scoring problems in Payday Loans. Nowadays, Payday Loans is experiencing its rapid development in China. The application of WHSBoost algorithm will further reduce the financial risk.

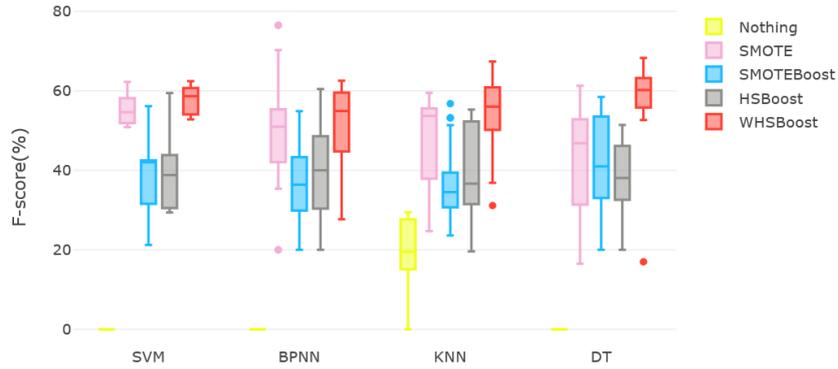

Fig.9 Box-plot of F-score on LC2018Q1Data data set

LC2018Q1Data data set has 107865 samples and 43 characteristics, and there are only 53 samples belonging to the minority class. As is shown in Fig.9, most of the algorithms are not ideal for classifying LC2018Q1Data data set. And the F-score of WHSBoost-SVM is 58.3009%, the highest in all models.

Fig.9 exhibits that the minimum and median of F-scores of WHSBoost algorithm are higher than the other algorithms, and the fluctuation is the smallest, and WHSBoost belongs to the first echelon. SMOTE have high accuracy, but there are great differences in different base classifiers. SMOTEBoost and HSBoost algorithm have the worst classification results and belong to the third echelons. For the SVM base classifier, the Recall of SMOTE-SVM is 87.485%, which are higher than that of WHSBoost-SVM. However, the F-score of SMOTE-SVM is 57.788%, lower than that of WHSBoost-SVM. That is to say, although SMOTE-SVM improves the accuracy of the minority class, it does not improve the overall accuracy. From the perspective of the principles of algorithms, SMOTE algorithm produces more minority samples by sampling. SMOTE algorithm may lead to a decrement in the accuracy of the majority class in the case of very few samples of the minority class. WHSBoost algorithm combines data sampling and ensemble algorithm and thus can effectively improve the overall accuracy. Therefore, on LC2018Q1Data data set, the F-score of WHSBoost algorithm is higher than the other algorithms.

In short, WHSBoost algorithm is still superior to the other algorithms in the case of large data imbalanced ratio. The validity and stability of WHSBoost algorithm on bank loan and Payday loan credit scoring problems are verified by experiments on 8 real credit-scoring data sets. At the same time, through data sets with different amounts of samples and different imbalanced ratios, we validate that the WHSBoost algorithm can solve the problem of classifying imbalanced data sets with different characteristics.

4.4.3 Non-parametric Test

In order to ensure that the result does not happen by accident, we test the significance between any two models. We use the non-parametric Wilcoxon signed-ranks test by F-scores of 50 repetitions in 8 credit scoring data sets, Australian, German, Japanese, Credit approval, DefaultData, QHaidata_A, QHaidata_B, LC2018Q1Data credit scoring data sets. The null hypothesis is that there is no significant difference between any two models. The alternative hypothesis is that the two models are significantly different. P value is significance of bilateral test.

We choose 0.05 as the confidence level. If the P value are less than 0.05, we will reject the null hypothesis and accept the alternative hypothesis. Table5-8 respectively shows the statistical significant test results of SVM, BPNN, KNN and DT combining different data processing algorithms. Each lattice represents a significant test of the P value between different algorithms and Z score between different algorithms. Z score is statistics of Wilcoxon matched-pairs signed ranks test. The red mark *indicates that the classification performance between the two models is significant. The mark 'b' indicates that Z score is based on the negative rank. The mark 'c' indicates that Z score is based on the positive rank.

Table 5 The Wilcoxon Signed Rank Test for SVM

|            | SMOTE | SMOTE Boost | HSBoost | WHSBoost |
|------------|-------|-------------|---------|----------|
| Nothing    | 0.000* <br> -14.196$^b$ | 0.000* <br> -17.184$^b$ | 0.000* <br> -17.027$^b$ | 0.000* <br> -17.332$^b$ |
| SMOTE      |       | 0.000* <br> -4.454$^c$ | 0.000* <br> -6.420$^c$ | 0.000* <br> -15.321$^b$ |
| SMOTEBoost |       |             | 0.026* <br> -2.222$^c$ | 0.000* <br> -17.186$^b$ |
| HSBoost    |       |             |         | 0.000* <br> -17.267$^b$ |

Table 6 The Wilcoxon Signed Rank Test for BPNN

|            | SMOTE | SMOTE Boost | HSBoost | WHSBoost |
|------------|-------|-------------|---------|----------|
| Nothing    | 0.000* <br> -13.977$^b$ | 0.000* <br> -14.327$^b$ | 0.000* <br> -13.887$^b$ | 0.000* <br> -15.683$^b$ |
| SMOTE      |       | 0.000* <br> -6.057$^c$ | 0.000* <br> -5.777$^c$ | 0.000* <br> -3.939$^b$ |
| SMOTEBoost |       |             | 0.948 <br> -0.66$^c$ | 0.000* <br> -11.084$^b$ |
| HSBoost    |       |             |         | 0.000* <br> -11.152$^b$ |

Table 7 The Wilcoxon Signed Rank Test for KNN

|            | SMOTE | SMOTE Boost | HSBoost | WHSBoost |
|------------|-------|-------------|---------|----------|
| Nothing    | 0.000* <br> -16.075$^b$ | 0.000* <br> -10.347$^b$ | 0.000* <br> -10.605$^b$ | 0.000* <br> -17.225$^b$ |
| SMOTE      |       | 0.000* <br> -13.187$^c$ | 0.000* <br> -10.984$^c$ | 0.000* <br> -11.505$^b$ |
| SMOTEBoost |       |             | 0.001* <br> -3.431$^b$ | 0.000* <br> -15.905$^b$ |
| HSBoost    |       |             |         | 0.000* <br> -15.065$^b$ |

Table 8 The Wilcoxon Signed Rank Test for DT

|         | SMOTE | SMOTE Boost | HSBoost | WHSBoost |
|---------|-------|-------------|---------|----------|
| Nothing | 0.000* | 0.000* | 0.000* | 0.000* |

|  | -13.476[b] | -13.045[b] | -13.194[b] | -15.347[b] |
|---|---|---|---|---|
| SMOTE |  | 0.021* | 0.000* | 0.000* |
|  |  | -3.080[c] | -9.097[c] | -11.060[b] |
| SMOTEBoost |  |  | 0.000* | 0.000* |
|  |  |  | -7.252[c] | -11.044[b] |
| HSBoost |  |  |  | 0.000* |
|  |  |  |  | -14.298[b] |

Tables 5-8 show that almost all types of data processing algorithms are effective for the SVM, BPNN, KNN and DT base classifiers. It demonstrates that base classifiers cannot effectively solve the imbalanced problems in classification, and thus it is necessary to propose different data processing algorithms to help the base classifier classify imbalanced data. Comparing the P values of different algorithms, we can see that the effects of SMOTEBoost and HSBoost algorithms are similar than others. On the BPNN base classifiers, the P values between HSBoost algorithm and SMOTEBoost algorithm is higher than 0.05 as well. SMOTEBoost and HSBoost algorithms are difficult to distinguish on BPNN. There are obvious differences between WHSBoost algorithm and SVM, BPNN, KNN, DT base classifiers. For the SVM, BPNN and KNN base classifiers, the P values between WHSBoost algorithm and SMOTE, HS, SMOTEBoost, RUSBoost and HSBoost are all around 0, which suggests distinct differences. In short, through statistical analysis test, WHSBoost has been verified to be effective in solving the problems of classifying imbalanced data sets.

## 5 Conclusions

The issue of bank credit-scoring is one of the most frequently encountered problems in banking business, and also a crucial part. Credit scoring problems can be regarded as a classifying issue on imbalanced data sets. We are supposed to ensure the recognition rate of "the bad credit samples", and at the same time, the misjudgment rate of "the good credit samples" should be reduced as much as possible. In addition, the credit-scoring model are supposed to have good stability and generalization ability. Therefore, we use the Recall, F-score and AUC as the evaluation criteria to find a better credit-scoring model.

In order to solve the problem of imbalanced data in classification, there are usually 3 kinds of model improvement methods: Feature selection, Data processing and Algorithm-based approaches. The ensemble algorithm based on Data processing is the most common improvement. In this paper, the problems existing in the general Adaboost algorithm based on Data Processing are analyzed, and a new ensemble algorithm, namely, WHSBoost, is proposed to deal with them. WHSBoost improves the weighted-based sampling process based on the adaboost algorithm, and solves the problem that the base classifier cannot process the weighted samples effectively in Adaboost, which guarantees the full use of the weight information and has a good effect on imbalanced sample data at the same time.

The main idea of WHSBoost is as follows. During each iteration of the ensemble algorithm, the WSMOTE algorithm and the WUSample algorithm process the weighted imbalanced data set with weights to obtain a temporary unweighted balanced data set, and then we can obtain the base classifier and the weights of the updated samples by training the balanced data set mentioned above. WHSBoost can not only improve the classification results, but also reduce the randomness of the algorithm, making the algorithm more stable.

In order to verify the applicability and robustness of the WHSBoost algorithm, we performed experiments on 20 simulation data sets with different imbalanced ratios and experiments on 8 real credit-scoring data sets respectively. The SMOTE, SMOTEBoost and HSBoost algorithms based on SVM, BPNN and KNN are constructed and we compare them with the WHSBoost algorithm.

It can be known from the simulations and experimental results of 20 real data sets with different imbalanced ratios:

- Comparing the performance of the 5 data processing algorithms on different imbalanced data sets, WHSBoost outperformed the others. With the change of the imbalanced ratios, the fluctuations of other algorithms are obviously larger than that of WHSBoost. Moreover, WHSBoost guarantees the high efficiency and stability of classification in different imbalanced ratios.
- Comparing the performance of different data processing methods on different base classifiers, the WHSBoost algorithm is superior to others on imbalanced data sets, and has the most obvious improvement effect on SVM.

Aiming at credit scoring problems, we conducted further experiments on 8 real credit scoring data sets. The results verified that WHSBoost-SVM is more appropriate than other algorithms.

- From the average of the overall production results, the Recall, F-score, AUC of WHSBoost are superior to other algorithms, belonging to the first echelon. SMOTE and SMOTEBoost are similar and belong to the second echelon. The performance of HSBoost is slightly worse and belongs to the third echelon.
- From the different data sets, the Recall, F-score, and AUC of WHSBoost are the best in most cases. What's more, the Japanese data set has better classification than the other data sets. In terms of QHaidata_B data set and LC2018Q1Data data set, this paper validates the validity and stability of WHSBoost algorithm in Payday loan credit scoring problems, which is pretty meaningful in economics.
- From the perspective of the base classifiers, the impact of imbalanced data on SVM is the greatest, and the impact on BPNN is the smallest. Meanwhile the base classifiers have some effects on the accuracy of credit scoring. Our experiments indicate that the classification result of WHSBoost-SVM on the same data set is better than WHSBoost-BPNN, WHSBoost-KNN in the credit scoring problems.

In conclusion, the experiments on simulation data sets and real data sets verify the validity of the WHSBoost proposed in this paper, and its applicability, robustness and good generalization ability are well reflected in imbalanced problems and credit scoring problems. WHSBoost can be well applied to real scenarios. Moreover, it can accurately identify "the bad samples" and reduce the misjudgment rate of "the good credit samples" in credit rating problems, which maximizes the economic benefits. However, the WHSBoost algorithm is supposed to have wider applications to more imbalanced problems in real life, which requires the further demonstration and improvement in the future work.

# References

Abdou, H. A., & Pointon, J. (2011). Credit scoring, statistical techniques and evaluation criteria: a review of the literature. Intelligent Systems in Accounting, Finance and Management, 18(2-3), 59-88.

Abellán, J., & Castellano, J. G. (2017). A comparative study on base classifiers in ensemble methods for credit scoring. Expert Systems with Applications, 73, 1-10.

Ala'raj, M., & Abbod, M. F. (2016). A new hybrid ensemble credit scoring model based on classifiers consensus system approach. Expert Systems with Applications, 64, 36-55.

Bae, S. H., & Yoon, K. J. (2015). Polyp detection via imbalanced learning and discriminative feature learning. IEEE transactions on medical imaging, 34(11), 2379-2393.

Bequé, A., & Lessmann, S. (2017). Extreme learning machines for credit scoring: An empirical evaluation. Expert Systems with Applications, 86, 42-53.

Bunkhumpornpat, C., Sinapiromsaran, K., & Lursinsap, C. (2009, April). Safe-level-smote: Safe-level-synthetic minority over-sampling technique for handling the class imbalanced problem. In Pacific-Asia conference on knowledge discovery and data mining (pp. 475-482). Springer, Berlin, Heidelberg.

Chawla, N. V., Bowyer, K. W., Hall, L. O., & Kegelmeyer, W. P. (2002). SMOTE: synthetic minority over-sampling technique. Journal of artificial intelligence research, 16, 321-357.

Chawla, N. V., Lazarevic, A., Hall, L. O., & Bowyer, K. W. (2003, September). SMOTEBoost: Improving prediction of the minority class in boosting. In European conference on principles of data mining and knowledge discovery (pp. 107-119). Springer, Berlin, Heidelberg.

Chen, C., Liaw, A., & Breiman, L. (2004). Using random forest to learn imbalanced data. University of California, Berkeley, 110, 1-12.

Cheng, F., Zhang, J., Wen, C., Liu, Z., & Li, Z. (2017). Large cost-sensitive margin distribution machine for imbalanced data classification. Neurocomputing, 224, 45-57.

D'Addabbo, A., & Maglietta, R. (2015). Parallel selective sampling method for imbalanced and large data classification. Pattern Recognition Letters, 62, 61-67.

Douzas, G., & Bacao, F. (2017). Self-Organizing Map Oversampling (SOMO) for imbalanced data set learning. Expert systems with Applications, 82, 40-52.

Galar, M., Fernández, A., Barrenechea, E., & Herrera, F. (2013). EUSBoost: Enhancing ensembles for highly imbalanced data-sets by evolutionary undersampling. Pattern Recognition, 46(12), 3460-3471.

García, V., Sánchez, J. S., & Mollineda, R. A. (2012). On the effectiveness of preprocessing methods when dealing with different levels of class imbalance. Knowledge-Based Systems, 25(1), 13-21.

Ghodselahi, A. (2011). A hybrid support vector machine ensemble model for credit scoring. International Journal of Computer Applications, 17(5), 1-5.

Gong, J., & Kim, H. (2017). RHSBoost: Improving classification performance in imbalance data. Computational Statistics & Data Analysis, 111, 1-13.

Haixiang, G., Yijing, L., Yanan, L., Xiao, L., & Jinling, L. (2016). BPSO-Adaboost-KNN ensemble learning algorithm for multi-class imbalanced data classification. Engineering Applications of Artificial Intelligence, 49, 176-193.

Haixiang, G., Yijing, L., Shang, J., Mingyun, G., Yuanyue, H., & Bing, G. (2017). Learning from class-imbalanced data: Review of methods and applications. Expert Systems with Applications, 73, 220-239.

Han, H., Wang, W. Y., & Mao, B. H. (2005, August). Borderline-SMOTE: a new over-sampling method in imbalanced data sets learning. In International Conference on Intelligent Computing (pp. 878-887). Springer, Berlin, Heidelberg.

Hart, P. (1968). The condensed nearest neighbor rule (Corresp.). IEEE transactions on information theory, 14(3), 515-516.

He, H., Zhang, W., & Zhang, S. (2018). A novel ensemble method for credit scoring: Adaption of different imbalanced ratios. Expert Systems with Applications, 98, 105-117.

Kubat, M., & Matwin, S. (1997, July). Addressing the curse of imbalanced training data sets: one-sided selection. In Icml (Vol. 97, pp. 179-186).

Laurikkala, J. (2001, July). Improving identification of difficult small classes by balancing class distribution. In Conference on Artificial Intelligence in Medicine in Europe (pp. 63-66). Springer, Berlin, Heidelberg.

Li, F., Zhang, X., Zhang, X., Du, C., Xu, Y., & Tian, Y. C. (2018). Cost-sensitive and hybrid-attribute measure multi-decision tree over imbalanced data sets. Information Sciences, 422, 242-256.

Lu, Zhai. (2016).Application of A New Model Based on Boosting Algorithm in Bank Credit Scoring Beijing Jiao tong University.

M'hamed, B. A., & Fergani, B. (2014). A new multi-class WSVM classification to imbalanced human activity data set. Journal of Computers, 9(7).

Maciejewski, T., & Stefanowski, J. (2011, April). Local neighbourhood extension of SMOTE for mining imbalanced data. In Computational Intelligence and Data Mining (CIDM), 2011 IEEE Symposium on (pp. 104-111). IEEE.

Maldonado, S., Weber, R., & Famili, F. (2014). Feature selection for high-dimensional class-imbalanced data sets using Support Vector Machines. Information Sciences, 286, 228-246.

Moayedikia, A., Ong, K. L., Boo, Y. L., Yeoh, W. G., & Jensen, R. (2017). Feature selection for high dimensional imbalanced class data using harmony search. Engineering Applications of Artificial Intelligence, 57, 38-49.

Roli, F. (2015). Multiple classifier systems. Encyclopedia of Biometrics, 1142-1147.

Seiffert, C., Khoshgoftaar, T. M., & Van Hulse, J. (2009). Hybrid sampling for imbalanced data. Integrated Computer-Aided Engineering, 16(3), 193-210.

Seiffert, C., Khoshgoftaar, T. M., Van Hulse, J., & Napolitano, A. (2010). RUSBoost: A hybrid approach to alleviating class imbalance. IEEE Transactions on Systems, Man, and Cybernetics-Part A: Systems and Humans, 40(1), 185-197.

Soleymani, R., Granger, E., & Fumera, G. (2018). Progressive boosting for class imbalance and its application to face re-identification. Expert Systems with Applications, 101, 271-291.

Tahir, M. A., Kittler, J., Mikolajczyk, K., & Yan, F. (2009, June). A multiple expert approach to the class imbalance problem using inverse random under sampling. In International Workshop on Multiple Classifier Systems (pp. 82-91). Springer, Berlin, Heidelberg.

Tomek, I. (1976). Two modifications of CNN. IEEE Trans. Systems, Man and Cybernetics, 6, 769-772.

Wang, Y., Liu, Y., Feng, L., & Zhu, X. (2015). Novel feature selection method based on harmony search for email classification. Knowledge-Based Systems, 73, 311-323.

Wilson, D. L. (1972). Asymptotic properties of nearest neighbor rules using edited data. IEEE Transactions on Systems, Man, and Cybernetics, (3), 408-421.

Xia, Y., Liu, C., Li, Y., & Liu, N. (2017). A boosted decision tree approach using Bayesian hyper-parameter optimization for credit scoring. Expert Systems with Applications, 78, 225-241.

Yin, L., Ge, Y., Xiao, K., Wang, X., & Quan, X. (2013). Feature selection for high-dimensional imbalanced data. Neurocomputing, 105, 3-11.

# Appendix I. Experimental Results on Data Sets with Different Imbalanced Ratios

Table 1 the mean of the numerical results on SVM

| algorithm | | Nothing | SMOTE | SMOTEBoost | HSBoost | WHSBoost |
|---|---|---|---|---|---|---|
| 45% minority class | Recall | 82.222 | 97.289 | 97.681 | 92.104 | **98.452** |
| | F-score | 85.426 | 94.372 | 94.059 | 90.062 | **95.770** |
| | AUC | 0.966 | 0.988 | 0.493 | 0.845 | **0.998** |
| 30% minority class | Recall | 48.000 | 96.222 | 96.189 | 88.867 | **96.600** |
| | F-score | 73.539 | 92.431 | 89.881 | 85.439 | **93.355** |
| | AUC | 0.962 | **0.997** | 0.652 | 0.899 | **0.997** |
| 20% minority class | Recall | 22.550 | 94.800 | 95.650 | 94.367 | **96.497** |
| | F-score | 64.392 | 94.366 | 92.085 | 86.064 | **95.112** |
| | AUC | 0.957 | 0.986 | 0.690 | 0.306 | **0.997** |
| 10% minority class | Recall | 93.277 | 93.367 | 90.433 | 81.067 | **93.767** |
| | F-score | 35.060 | 91.807 | 90.322 | 73.334 | **93.178** |
| | AUC | 0.664 | 0.993 | 0.979 | 0.892 | **0.995** |

Table 2 the mean of the numerical results on BPNN

| algorithm | | Nothing | SMOTE | SMOTEBoost | HSBoost | WHSBoost |
|---|---|---|---|---|---|---|
| 45% minority class | Recall | 96.000 | 97.148 | 96.407 | 95.815 | **98.074** |
| | F-score | 95.573 | 96.562 | 96.173 | 95.800 | **97.864** |
| | AUC | 0.975 | 0.969 | 0.967 | 0.961 | **0.982** |
| 30% minority class | Recall | 95.333 | 97.389 | 94.111 | 94.167 | **97.444** |
| | F-score | 95.550 | 96.458 | 94.451 | 91.822 | **96.502** |
| | AUC | 0.957 | 0.969 | 0.966 | 0.947 | **0.973** |
| 20% minority class | Recall | 88.483 | 95.833 | 94.000 | 89.117 | **96.333** |
| | F-score | 88.091 | 92.992 | 92.851 | 87.754 | **96.185** |
| | AUC | 0.942 | 0.963 | 0.963 | 0.930 | **0.975** |
| 10% minority class | Recall | 84.500 | 92.000 | 90.167 | 82.500 | **94.667** |
| | F-score | 86.116 | 90.459 | 88.274 | 79.947 | **92.320** |
| | AUC | 0.950 | 0.948 | 0.952 | 0.907 | **0.969** |

Table 3 the mean of the numerical results on KNN

| algorithm | | Nothing | SMOTE | SMOTEBoost | HSBoost | WHSBoost |
|---|---|---|---|---|---|---|
| 45% minority class | Recall | 69.222 | 78.119 | 75.400 | 76.474 | **92.733** |
| | F-score | 69.468 | 78.893 | 73.998 | 70.450 | **88.532** |
| | AUC | **0.823** | 0.809 | 0.807 | 0.810 | 0.820 |
| 30% minority class | Recall | 44.100 | 82.744 | 59.556 | 57.478 | **87.733** |
| | F-score | 41.555 | 80.560 | 67.671 | 65.531 | **83.163** |
| | AUC | 0.811 | 0.828 | 0.811 | 0.829 | **0.839** |
| 20% minority class | Recall | 24.283 | 78.433 | 61.567 | 63.717 | **82.733** |
| | F-score | 25.704 | 72.369 | 62.760 | 59.818 | **73.449** |
|  | AUC | 0.780 | 0.769 | 0.806 | 0.827 | **0.829** |
|  | Recall | 6.867 | 78.533 | 60.433 | 63.567 | **79.900** |
| 10% minority class | F-score | 6.934 | 61.191 | 50.607 | 50.789 | **64.628** |
|  | AUC | 0.703 | 0.757 | 0.799 | **0.821** | 0.819 |

Table 4 the mean of the numerical results on DT

| algorithm |  | Nothing | SMOTE | SMOTEBoost | HSBoost | WHSBoost |
|---|---|---|---|---|---|---|
|  | Recall | 56.611 | 82.844 | 67.107 | 61.148 | **90.446** |
| 45% minority class | F-score | 60.780 | 82.453 | 66.932 | 62.370 | **84.365** |
|  | AUC | 0.675 | 0.779 | 0.684 | 0.672 | **0.808** |
|  | Recall | 45.500 | **72.778** | 65.167 | 63.722 | 72.056 |
| 30% minority class | F-score | 55.002 | 72.291 | 66.325 | 59.744 | **73.584** |
|  | AUC | 0.672 | 0.748 | 0.712 | 0.710 | **0.793** |
|  | Recall | 48.167 | 65.167 | 61.000 | 58.167 | **69.167** |
| 20% minority class | F-score | 48.802 | 61.199 | 57.017 | 53.165 | **63.887** |
|  | AUC | 0.694 | 0.729 | 0.726 | 0.726 | **0.750** |
|  | Recall | 3.167 | 60.667 | 41.833 | 32.667 | **65.667** |
| 10% minority class | F-score | 3.074 | 51.995 | 47.529 | 37.502 | **54.158** |
|  | AUC | 0.514 | 0.693 | 0.680 | 0.657 | **0.700** |

Table 5 The 20 real data sets

| Name | N | Features | IR | Name | N | Features | IR |
|---|---|---|---|---|---|---|---|
| ecoli-0_vs_1 | 220 | 7 | 1.86 | pima | 768 | 8 | 1.9 |
| ecoli2 | 336 | 7 | 5.46 | Segmentation0 | 2308 | 19 | 6.01 |
| ecoli3 | 336 | 7 | 8.19 | Vehicle0 | 846 | 18 | 3.23 |
| glass-0-1-2-3_vs_4-5-6 | 214 | 9 | 3.19 | Vehicle1 | 846 | 18 | 2.52 |
| glass0 | 214 | 9 | 2.06 | Vehicle2 | 846 | 18 | 2.52 |
| glass6 | 214 | 9 | 6.38 | wisconsin | 683 | 9 | 1.86 |
| Haberman | 306 | 3 | 2.68 | yeast3 | 1484 | 8 | 8.11 |
| new-thyroid2 | 215 | 5 | 4.92 | ecoli-0-1-3-7_vs_2-6 | 281 | 7 | 39.15 |
| page-blocks0 | 5472 | 10 | 8.77 | ecoli4 | 336 | 7 | 13.84 |
| ecoli-0_vs_1 | 220 | 7 | 1.86 | page-blocks-1-3_vs_4 | 472 | 10 | 15.85 |

*In Table 5, "N" is the number of samples in the data sets. Column "Features" is the number of features. "Column "IR refers to the proportion of positive samples and negative samples, and we generally define the minority as the positive class. In order to verify the classification effect, we selected 80% of credit scoring data sets randomly as the training samples, and the rest as the testing samples.

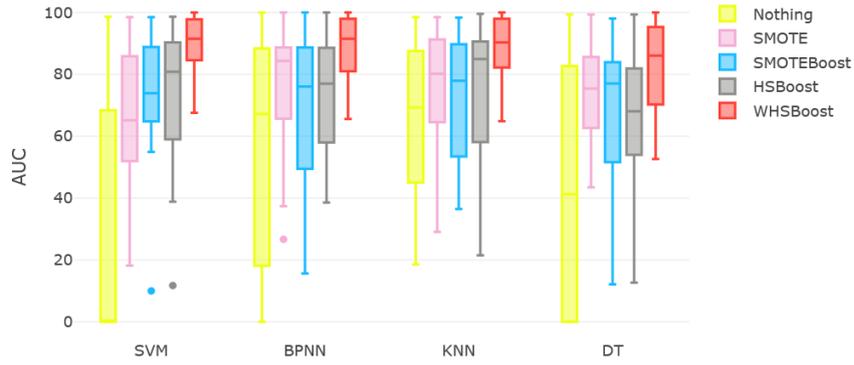

Fig.1 AUC of 20 Real Data sets

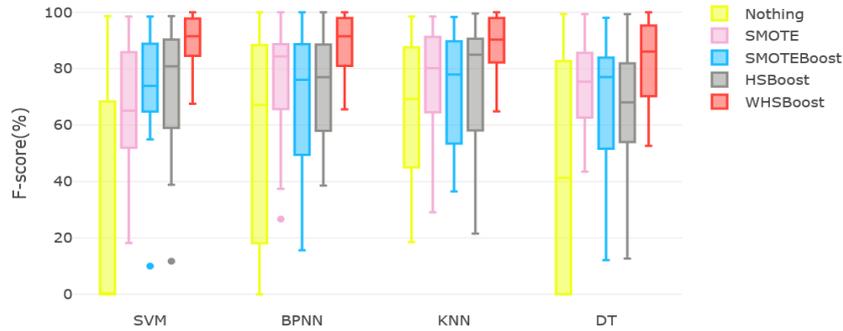

Fig.2 F-score of 20 Real Data sets

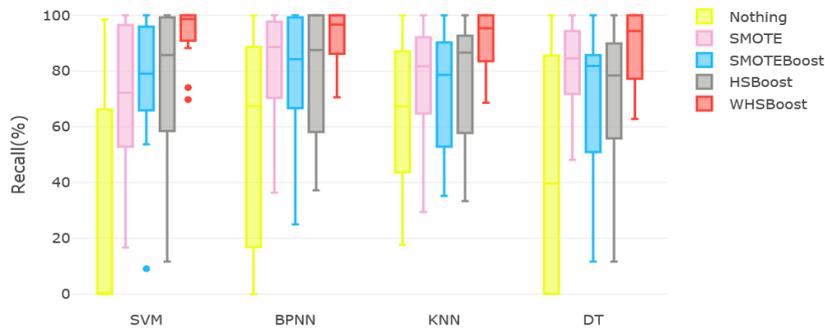

Fig.3 F-score of 20 Real Data sets

## AppendixⅡ. Experimental Results of credit scoring data sets

Table 1 F-score of classification on credit scoring data sets

|     |            | German | Australian | Japanese | Credit approval | Default Data | QHaidata_A | QHaidata_B | LC2018Q1Data |
|-----|------------|--------|------------|----------|-----------------|--------------|------------|------------|--------------|
| SVM | Nothing    | 0.000  | 76.824     | 57.092   | 0.000           | 0.000        | 0.000      | 0.000      | 0.000        |
|     | SMOTE      | 79.171 | 80.216     | 65.674   | 58.668          | 76.376       | 71.872     | 63.250     | 55.410       |
|     | SMOTEBoost | 48.577 | 82.579     | 81.946   | 67.536          | 55.877       | 46.368     | 60.983     | 38.651       |
|     | HSBoost    | 50.184 | 81.491     | 80.965   | 49.819          | 59.218       | 46.603     | 53.813     | 39.340       |
|     | WHSBoost   | **82.010** | **91.346** | **91.146** | **79.287**  | **79.746**   | **72.265** | **73.199** | **57.788**   |
| BPNN| Nothing    | 44.168 | 84.882     | 82.106   | 0.971           | 5.136        | 0.861      | 0.000      | 0.000        |
|     | SMOTE      | 58.030 | 82.801     | 82.925   | 57.528          | 61.630       | 60.715     | 66.020     | 50.328       |

|  |  |  |  |  |  |  |  |  |
|---|---|---|---|---|---|---|---|---|
|  | SMOTEBoost | 59.959 | 84.190 | 85.257 | 42.324 | 45.477 | 54.539 | 52.696 | 37.047 |
|  | HSBoost | 60.854 | 84.753 | 83.695 | 49.152 | 48.124 | 53.537 | 52.642 | 39.103 |
|  | WHSBoost | **66.900** | **87.292** | **87.896** | **61.754** | **60.478** | **66.345** | **70.198** | **51.109** |
| KNN | Nothing | 44.822 | 84.269 | 85.763 | 7.197 | 17.649 | 1.719 | 13.800 | 18.085 |
|  | SMOTE | 65.939 | 85.507 | 86.484 | 62.154 | 58.154 | 49.341 | 55.751 | 46.657 |
|  | SMOTEBoost | 57.288 | 78.712 | 77.542 | 54.770 | 55.070 | 42.947 | 50.361 | 36.984 |
|  | HSBoost | 60.549 | 78.677 | 77.818 | 61.384 | 60.166 | 48.129 | 45.486 | 38.583 |
|  | WHSBoost | **72.907** | **88.322** | **88.140** | **74.992** | **63.771** | **52.230** | **65.391** | **54.339** |
| DT | Nothing | 0.000 | 71.478 | 67.664 | 0.000 | 0.000 | 0.000 | 0.000 | 0.000 |
|  | SMOTE | 63.501 | **71.539** | 65.623 | 51.290 | 52.038 | 58.316 | 58.843 | 42.641 |
|  | SMOTEBoost | 60.172 | 68.959 | 65.183 | 42.869 | 56.668 | 62.278 | 60.907 | 41.518 |
|  | HSBoost | 28.760 | 69.010 | 64.717 | 40.358 | 51.954 | 50.671 | 33.375 | 37.690 |
|  | WHSBoost | **63.853** | 70.972 | **71.969** | **62.002** | **62.684** | **72.231** | **78.452** | **58.301** |

Table 2 Recall of classification on credit scoring data sets

|  |  | German | Australian | Japanese | Credit approval | Default Data | QHaidata_A | QHaidata_B | LC2018 Q1Data |
|---|---|---|---|---|---|---|---|---|---|
| SVM | Nothing | 0.000 | 74.918 | 54.973 | 0.000 | 0.000 | 0.000 | 0.000 | 0.000 |
|  | SMOTE | 86.833 | 80.885 | 63.959 | 59.017 | 85.200 | 83.684 | 70.000 | 87.485 |
|  | SMOTEBoost | 48.417 | 84.262 | 83.824 | 68.548 | 38.200 | 60.000 | 64.583 | 32.632 |
|  | HSBoost | 50.583 | 83.639 | 79.730 | 46.843 | 54.400 | 43.158 | 51.250 | 32.368 |
|  | WHSBoost | **92.167** | **93.279** | **98.284** | **82.983** | **90.600** | **84.737** | **90.417** | **85.417** |
| BPNN | Nothing | 43.917 | 85.770 | 83.338 | 0.909 | 6.400 | 0.789 | 0.000 | 0.000 |
|  | SMOTE | 59.917 | 84.230 | 84.446 | 56.462 | 44.800 | 48.421 | 55.167 | 45.000 |
|  | SMOTEBoost | 61.083 | 87.279 | 88.162 | 33.627 | 26.600 | 38.684 | 25.833 | 24.167 |
|  | HSBoost | 60.917 | 87.131 | 84.824 | 43.550 | 30.600 | 27.368 | 31.250 | 31.667 |
|  | WHSBoost | **67.667** | **89.393** | **90.581** | **66.719** | **54.000** | **48.421** | **61.667** | **49.667** |
| KNN | Nothing | 42.083 | 74.574 | 86.000 | 6.769 | 17.800 | 1.579 | 22.083 | 21.842 |
|  | SMOTE | 68.667 | 86.164 | 86.905 | 54.353 | 46.600 | 43.263 | 45.833 | 53.684 |
|  | SMOTEBoost | 58.667 | 86.311 | 77.527 | 42.454 | 44.200 | 24.737 | 31.250 | 29.211 |
|  | HSBoost | 62.167 | 86.180 | 77.811 | 46.405 | 48.400 | 41.316 | 27.917 | 42.895 |
|  | WHSBoost | **78.917** | **92.918** | **89.324** | **79.972** | **56.600** | **48.211** | **67.083** | **71.000** |
| DT | Nothing | 0.000 | 73.049 | 61.203 | 0.000 | 0.000 | 0.000 | 0.000 | 0.000 |
|  | SMOTE | 67.500 | **73.131** | 72.689 | 54.330 | 48.600 | 18.684 | 50.833 | 36.667 |
|  | SMOTEBoost | 58.667 | 68.016 | 70.081 | 43.321 | 46.000 | 22.632 | 24.583 | 25.833 |
|  | HSBoost | 32.583 | 70.016 | 69.608 | 28.031 | 40.600 | 47.632 | 26.667 | 24.583 |
|  | WHSBoost | **69.667** | 69.770 | **73.135** | **66.397** | **51.200** | **55.263** | **85.833** | **69.583** |

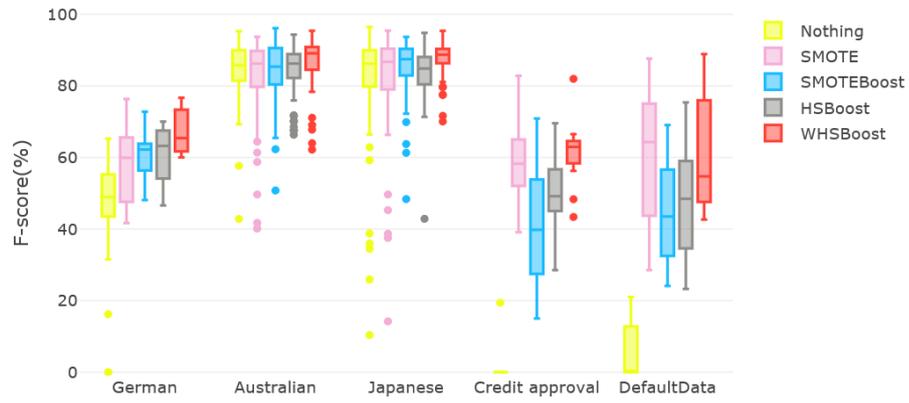

Fig.1 F-score of Australian, German, Japanese, Credit approval, DefaultData data sets on BPNN

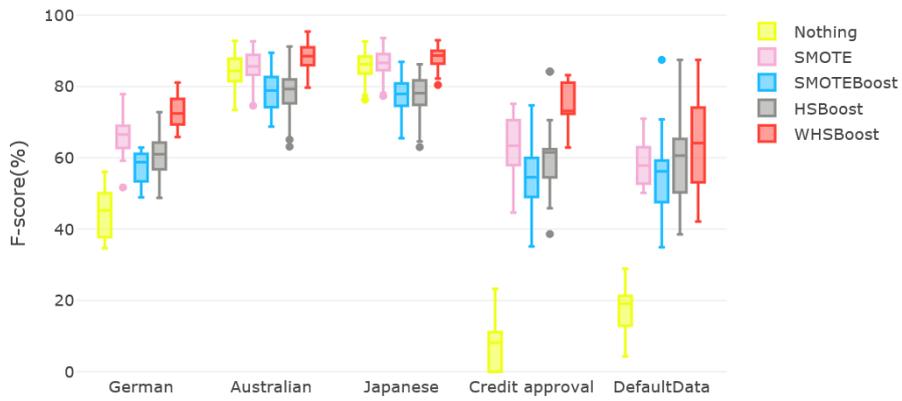

Fig.2 F-score of Australian, German, Japanese, Credit approval, DefaultData data sets on KNN

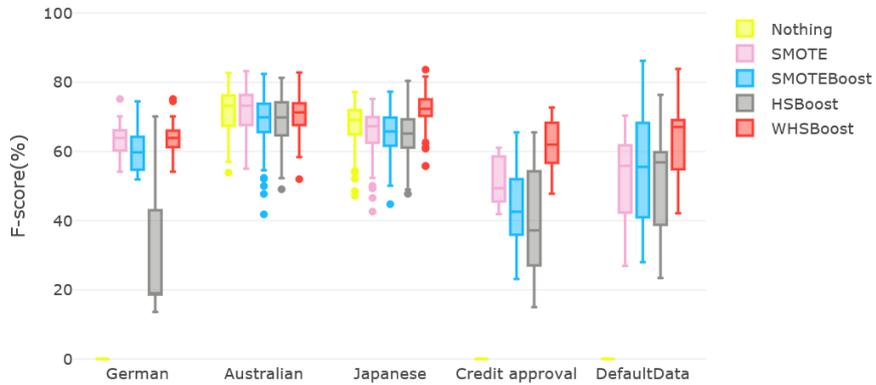

Fig.3 F-score of Australian, German, Japanese, Credit approval, DefaultData data sets on DT